%% file: 0_main.tex
\documentclass[11pt,a4paper]{article}
\usepackage[hyperref]{emnlp2018}
\usepackage{times}
\usepackage{latexsym}
\usepackage{color}

\usepackage{url}

\usepackage{helvet}  
\usepackage{courier}  
\usepackage{graphicx}  
\usepackage{multirow}
\usepackage{amsmath, amsfonts, amssymb, xspace, color,listings,enumitem,graphicx}
\usepackage{natbib}
\usepackage{diagbox}

\usepackage{moresize}

\usepackage{tabularx, booktabs}
\newcolumntype{Y}{>{\centering\arraybackslash}X}

\def \our {\mbox{LD-Net}\xspace}
\def \plm {\mbox{LMs}\xspace}
\def \oel {\mbox{O-ELMo}\xspace}
\def \nel {\mbox{R-ELMo}\xspace}
\def \nolm {\mbox{NoLM}\xspace}

\def \v {\mathbf{v}}
\def \w {\mathbf{w}}
\def \x {\mathbf{x}}
\def \y {\mathbf{y}}
\def \z {\mathbf{z}}

\def \r {\mathbf{r}}
\def \c {\mathbf{c}}
\def \h {\mathbf{h}}
\def \u {\mathbf{u}}
\def \b {\mathbf{b}}
\def \c {\mathbf{c}}
\def \Y {\mathbf{Y}}
\def \U {\mathbf{U}}

\aclfinalcopy 

\title{Efficient Contextualized Representation:\\ Language Model Pruning for Sequence Labeling}

\author{
Liyuan Liu $^{\dag}\ \ $
Xiang Ren $^{\sharp}\ \ $
Jingbo Shang $^{\dag}\ \ $
Xiaotao Gu $^{\dag}\ \ $
Jian Peng $^{\dag}\ \ $
Jiawei Han $^{\dag}$
\\[0.5ex]
{$^{\dag}$ University of Illinois at Urbana-Champaign, Urbana, IL, USA}\\
{$^{\sharp}$ University of Southern California, Los Angeles, CA, USA}\\
{
\begin{small}
\tt
$^{\dag}$\{ll2, shang7, xiaotao2, jianpeng, hanj\}@illinois.edu $^{\sharp}$xiangren@usc.edu
\end{small}
}
}

\date{}

\begin{document}
\maketitle

\input{0abs}

\begin{figure}[t]
  \centering
    \includegraphics[width=\columnwidth]{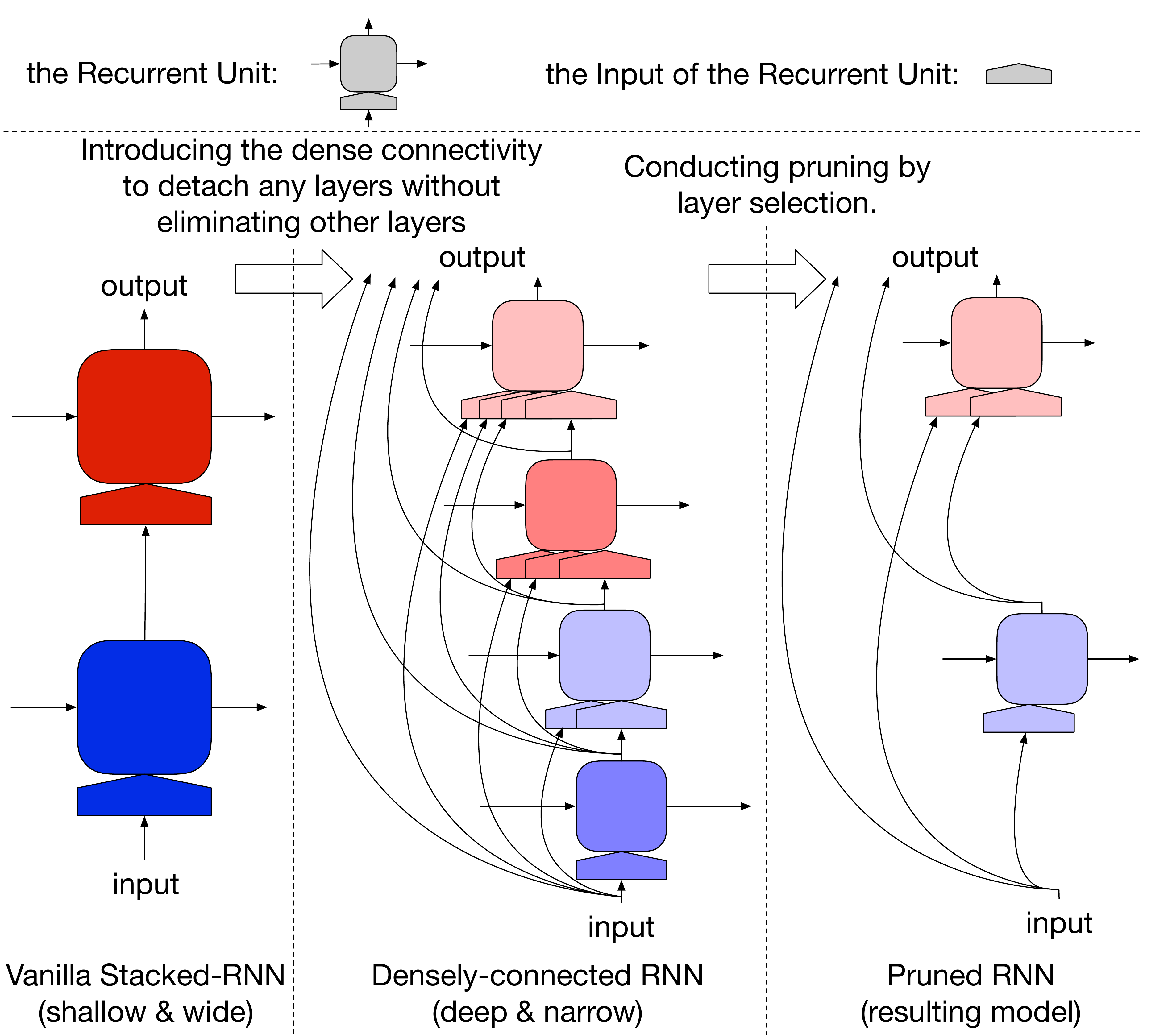}
  \caption{Leverage the dense connectivity to compress models via layer selection, and replace wide and shallow RNNs with deep and narrow ones.}
  \vspace{-0.3cm}
  \label{fig:Framework}
\end{figure}

\input{1intro}
\input{2method}
\input{3exp}
\input{4related}
\input{5con}
\input{6akn}

\newpage
\bibliography{cited}
\bibliographystyle{acl_natbib_nourl}

\end{document}

%% file: 0abs.tex

\begin{abstract}
Many efforts have been made to facilitate natural language processing tasks with pre-trained language models (\plm), and brought significant improvements to various applications.
To fully leverage the nearly unlimited corpora and capture linguistic information of multifarious levels, large-size LMs are required;
but for a specific task, only parts of these information are useful.
Such large-sized \plm, even in the inference stage, may cause heavy computation workloads,
making them too time-consuming for large-scale applications.
Here we propose to compress bulky \plm while preserving useful information 
with regard
to a specific task.
As different layers of the model keep different information, we develop a \textit{layer selection} method for model pruning using sparsity-inducing regularization.
By introducing the dense connectivity, we can detach any layer without affecting others, and stretch shallow and wide LMs to be \textit{deep} and \textit{narrow}.
In model training, \plm are learned with \textit{layer-wise dropouts} for better robustness.
Experiments on 
two benchmark datasets
demonstrate the effectiveness of our method.
\end{abstract}

%% file: 1intro.tex

\section{Introduction}

Benefited from the recent advances in neural networks (NNs) and the access to nearly unlimited corpora, 
neural language models are able to 
achieve a good perplexity score and 
generate high-quality sentences.
These LMs automatically capture abundant linguistic information and patterns from large text corpora, and can be applied to facilitate a wide range of NLP applications~\cite{rei2017semi,liu2017empower,e2018deep}.

Recently, efforts
have been made on learning contextualized representations with pre-trained language models (\plm) ~\cite{e2018deep}.
These pre-trained layers brought significant improvements to various NLP benchmarks, yielding up to $30\%$ relative error reductions.
However, due to high variability of language, 
gigantic NNs (e.g., LSTMs with 8,192 hidden states) are preferred to construct informative \plm and extract multifarious linguistic information~\cite{peters2017semi}.
Even though these models can be integrated without retraining (using their forward pass only), 
they still result in heavy computation workloads during inference stage, 
making them prohibitive for real-world applications.


In this paper, we aim to compress \plm for the end task in a plug-in-and-play manner.
Typically, NN compression methods require the retraining of the whole model~\cite{mellempudi2017ternary}.
However, neural language models are usually composed of RNNs, and their backpropagations require significantly more RAM than their inference.
It would become even more cumbersome when the target task equips the coupled \plm to capture information in both directions.
Therefore, these 
methods do not fit our scenario very well.
Accordingly, we try to compress \plm while \emph{avoiding costly retraining}.

Intuitively,
layers of different depths 
would
 capture linguistic information of different levels.
Meanwhile, since \plm are trained in a task-agnostic manner, not all layers and their extracted information are relevant to the end task.
Hence, we propose to 
compress the model by layer selection,
which retains useful layers for the target task and prunes irrelevant ones.
However, for the widely-used stacked-LSTM, directly pruning any layers will eliminate all subsequent ones.
To overcome this challenge, we introduce the dense connectivity.
As shown in Fig.~\ref{fig:Framework}, it allows us to detach any layers while keeping all remaining ones, thus 
creating the basis to avoid retraining.
Moreover, such connectivity can stretch shallow and wide LMs to be deep and narrow~\cite{huang2016densely}, and enable a more fine-grained layer selection.

Furthermore, we try to 
retain the effectiveness of the pruned model. 
Specifically, we modify the $L_1$ regularization 
for encouraging
the selection weights to be not only sparse but binary, 
which protects the retained layer connections from shrinkage.
Besides, we design a layer-wise dropout to make \plm more robust and better prepared for the layer selection.

We refer to our model as \our,
since the \emph{layer selection} and the \emph{dense connectivity} form the basis of our pruning methods.
For evaluation, 
we apply \our on two sequence labeling benchmark datasets, and demonstrated the effectiveness of the proposed method.
In the CoNLL03 Named Entity Recognition (NER) task, the $F_1$ score increases from 90.78$\pm$0.24\% to 91.86$\pm$0.15\% by integrating the unpruned \plm.
Meanwhile, after pruning over 90\% calculation workloads from the best performing model\footnote{Based on their performance on the development sets} (92.03\%), the resulting model still yields 91.84$\pm$0.14\%. Our implementations and pre-trained models would be released for futher study\footnote{
\url{https://github.com/LiyuanLucasLiu/LD-Net}.}.

%% file: 2method.tex

\section{\our}

Given a input sequence of $T$ word-level tokens, $\{x_1$, $x_2$, $\cdots$, $x_T\}$, we use $\x_t$ to denote the embedding 
of $x_t$.
For a $L$-layers NN, we mark the input and output of the $l^{th}$ layer at the $t^{th}$ time stamp as $\x_{l, t}$ and $\h_{l, t}$.

\subsection{RNN and Dense Connectivity}

We represent one RNN layer as a function:
\begin{equation}
\h_{l, t} = F_l(\x_{l, t}, \h_{l, t-1})
\label{eqn:v_rnn}
\end{equation}
where $F_l$ is the recurrent unit of $l^{th}$ layer, it could be 
any RNNs variants,
and 
the vanilla LSTMs
 is used 
in our experiments.

As deeper NNs usually have more representation power, RNN layers are often stacked together to form the final model by setting $\x_{l, t} = \h_{l - 1, t}$.
These vanilla stacked-RNN models, however, suffer from problems like the vanishing gradient, and it's hard to train very deep models.

Recently, the dense connectivity and residual connectivity have been proposed to handle these problems~\cite{he2016identity,huang2016densely}.
Specifically, dense connectivity refers to adding direct connections from any layer to all its subsequent layers.
As illustrated in Figure~\ref{fig:Framework}, the input of $l^{th}$ layer is composed of 
the original input and
the output of all preceding layers as follows.
$$
\x_{l, t} = [\x_t, \h_{1, t}, \cdots, \h_{l - 1, t}]
$$
Similarly, the final output of the $L$-layer RNN is $\h_t = [\x_t, \h_{1, t}, \cdots, \h_{L, t}]$.
With dense connectivity, we can detach any single layer without eliminating its subsequent layers (as in Fig.~\ref{fig:Framework}).
Also, existing practices in computer vision demonstrate that such connectivities can lead to deep and narrow NNs and distribute parameters into different layers.
Moreover, different layers in \plm usually capture linguistic information of different levels.
Hence, we can compress \plm for a specific task by pruning unrelated or unimportant layers. 

\subsection{Language Modeling}

Language modeling aims to describe the sequence generation.
Normally, the generation probability of the sequence $\{x_1, \cdots, x_T\}$ is defined in a ``forward'' manner:
\begin{equation}
p(x_1, \cdots, x_T) = \prod_{t=1}^T p(x_t | x_1, \cdots, x_{t-1})
\label{eqn:forw_lm}
\end{equation}
Where $p(x_t | x_1, \cdots, x_{t-1})$ is computed based on the output of RNN, $\h_t$.
Due to the dense connectivity, $\h_t$ is composed of 
outputs 
from different layers, which are designed to capture 
linguistic information of different levels.
Similar to the bottleneck layers employed in the DenseNet~\cite{huang2016densely}, we 
add
additional layers to unify such information.
Accordingly, we add an projection layer with the $\mathrm{ReLU}$ activation function:
\begin{equation}
\h_t^* = \mathrm{ReLU}(W_{proj} \cdot \h_t + \b_{proj})
\label{eqn:h_proj}
\end{equation}
Based on $\h_t^*$, it's intuitive to calculate $p(x_t | x_1, \cdots, x_{t-1})$ by the softmax function, i.e., $\mathrm{softmax}(W_{out} \cdot \h_t^* + \b)$.

Since the training of language models needs nothing but the raw text,
it has almost unlimited corpora.
However, 
conducting training
on extensive corpora results in a huge dictionary, 
and makes calculating the vanilla softmax intractable.
Several techniques have been proposed to handle this problem, including adaptive softmax~\cite{grave2016efficient}, slim word embedding~\cite{li2017slim}, the sampled softmax and the noise contrastive estimation~\cite{jozefowicz2016exploring}.
Since the major focus of our paper does not lie in the language modeling task, 
we choose the adaptive softmax because of its practical efficiency when accelerated with GPUs.

\subsection{Contextualized Representations}

As pre-trained LMs can describe the text generation accurately, they can be utilized to extract information and construct features for other tasks.
These features, referred as contextualized representations, have been demonstrated to be essentially useful~\cite{e2018deep}.
To capture information from both directions, we utilized not only forward LMs, but also backward LMs. 
Backward LMs are based on Eqn.~\ref{eqn:back_lm} instead of Eqn.~\ref{eqn:forw_lm}.
Similar to forward LMs, backward LMs approach $p(x_t | x_{t+1}, \cdots, x_T)$ with NNs.
For reference, the output of the RNN in backward LMs for $x_t$ is recorded as $\h_t^{r}$.
\begin{equation}
p(x_1, \cdots, x_n) = \prod_{t=1}^T p(x_t | x_{t+1}, \cdots, x_T)
\label{eqn:back_lm}
\end{equation}

Ideally, the final output of \plm (e.g., $\h_t^*$) would be the same as the representation of the target word (e.g., $x_{t+1}$); therefore, it may not contain much context information.
Meanwhile, the output of the densely connected RNN (e.g., $\h_t$) includes outputs from every layer, thus
summarizing all extracted features.
Since the dimensions of $\h_t$ could be too large for the end task, we add a non-linear transformation to calculate the contextualized representation ($\r_t$):
\begin{equation}
\r_t = \mathrm{ReLU}(W_{cr} \cdot [\h_t, \h_t^{r}] + \b_{cr})
\label{eqn:cr}
\end{equation}

Our proposed method bears the same intuition as the ELMo~\cite{e2018deep}.
ELMo is designed for the vanilla stacked-RNN, and tries to calculate a weighted average of different layers' outputs as the contextualized representation.
Our method, benefited from the dense connectivity and 
its narrow structure,
can directly combine the outputs of different layers by concatenation.
It does not assume the outputs of different layers to be in the same vector space, thus having more potential for transferring the constructed token representations.
More discussions are available in Sec.~\ref{sec:exp}.

\begin{figure*}[ht!]
  \centering
    \includegraphics[width=\textwidth]{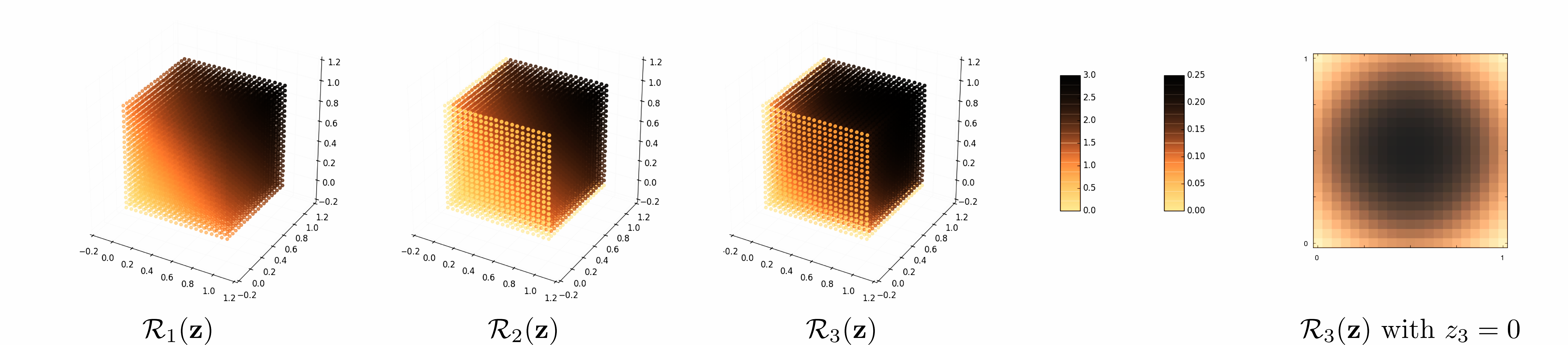}
\vspace{-0.8cm}
  \caption{Penalty values of various $\mathcal{R}$ for $\z$ with three dimensions. $\lambda_1$ has been set to $2$ for $\mathcal{R}_2$ and $\mathcal{R}_3$.}
  \label{fig:Reg}
\vspace{-0.5cm}
\end{figure*}

\subsection{Layer Selection}

Typical model compression methods require retraining or gradient calculation. 
For the coupled \plm,
these methods require even more computation resources compared to the training of \plm, thus not fitting our scenario very well.

Benefited from the dense connectivity,
we are able to train deep and narrow networks.
Moreover, we can detach one of its layer without eliminating all subsequent layers (as in Fig.~\ref{fig:Framework}).
Since different layers in NNs could capture different linguistic information, only a few of them would be relevant or useful for a specific task.
As a result, 
we try to 
compress these models
by the task-guided layer selection.
For $i$-th layer, we introduce a binary mask $z_i \in \{0, 1\}$ and calculate $\h_{l, t}$ with Eqn.~\ref{eqn:z_rnn} instead of Eqn.~\ref{eqn:v_rnn}.
\begin{equation}
\h_{l, t} = z_i \cdot F_l(\x_{l, t}, \h_{l, t-1})
\label{eqn:z_rnn}
\end{equation}
With this setting, we can conduct a layer selection by optimizing the regularized empirical risk:
\begin{equation}
\min \mathcal{L} + \lambda_0 \cdot \mathcal{R}
\label{eqn:emp_loss}
\end{equation}
where $\mathcal{L}$ is the empirical risk for the sequence labeling task and $\mathcal{R}$ is the sparse regularization.

The ideal choice for $\mathcal{R}$ would be the $L_0$ regularization of $\z$, i.e., $\mathcal{R}_0(\z) = |\z|_0$.
However, it is not continuous and cannot be efficiently optimized. 
Hence, we relax $z_i$ from binary to a real value (i.e., $0 \le z_i \le 1$) and replace $\mathcal{R}_0$ by:
$$
\mathcal{R}_1 = |\z|_1
$$

Despite the sparsity achieved by $\mathcal{R}_1$, it could hurt the performance by shifting all $z_i$ far away from $1$.
Such shrinkage introduces additional noise in $\h_{l, t}$ and $\x_{l, t}$, which may 
result in ineffective pruned \plm.
Since our goal is to conduct pruning without retraining, we further modify the $L_1$ regularization to achieve sparsity while alleviating its shrinkage effect.
As the target of $\mathcal{R}$ is to make $\z$ sparse, it can be ``turned-off'' after achieving a satisfying sparsity.
Therefore, we extend $\mathcal{R}_1$ to a margin-based regularization:
$$
\mathcal{R}_2 = \delta(|\z|_0 > \lambda_1) |\z|_1
$$

In addition, 
we also want to make up the relaxation made on $\z$, i.e., relaxing its values from binary to $[0, 1]$.
Accordingly, we add the penalty $|\z (1 - \z)|_1$ to encourage $\z$ to be binary~\cite{murray2010algorithm} and modify $\mathcal{R}_2$ into $\mathcal{R}_3$:
$$
\mathcal{R}_3= \delta(|\z|_0 > \lambda_1) |\z|_1 + |\z (1-\z)|_1
$$

To compare $\mathcal{R}_1$, $\mathcal{R}_2$ and $\mathcal{R}_3$, we visualize their penalty values in Fig.~\ref{fig:Reg}.
The visualization is generated for a $3$-dimensional $\z$ while the targeted sparsity, $\lambda_1$, is set to $2$.
Comparing to $\mathcal{R}_1$, we can observe that $\mathcal{R}_2$ enlarges the optimal point set from $\mathbf{0}$ to all $\z$ with a satisfying sparsity, thus avoiding the over-shrinkage.
To better demonstrate the effect of $\mathcal{R}_3$, 
we further visualize its penalties after achieving a satisfying sparsity (w.l.o.g., assuming $z_3 = 0$).
One can observe that it penalizes non-binary $\z$ and favors binary values. 

\subsection{Layer-wise Dropout}

\begin{figure}
  \centering
    \includegraphics[width=\columnwidth]{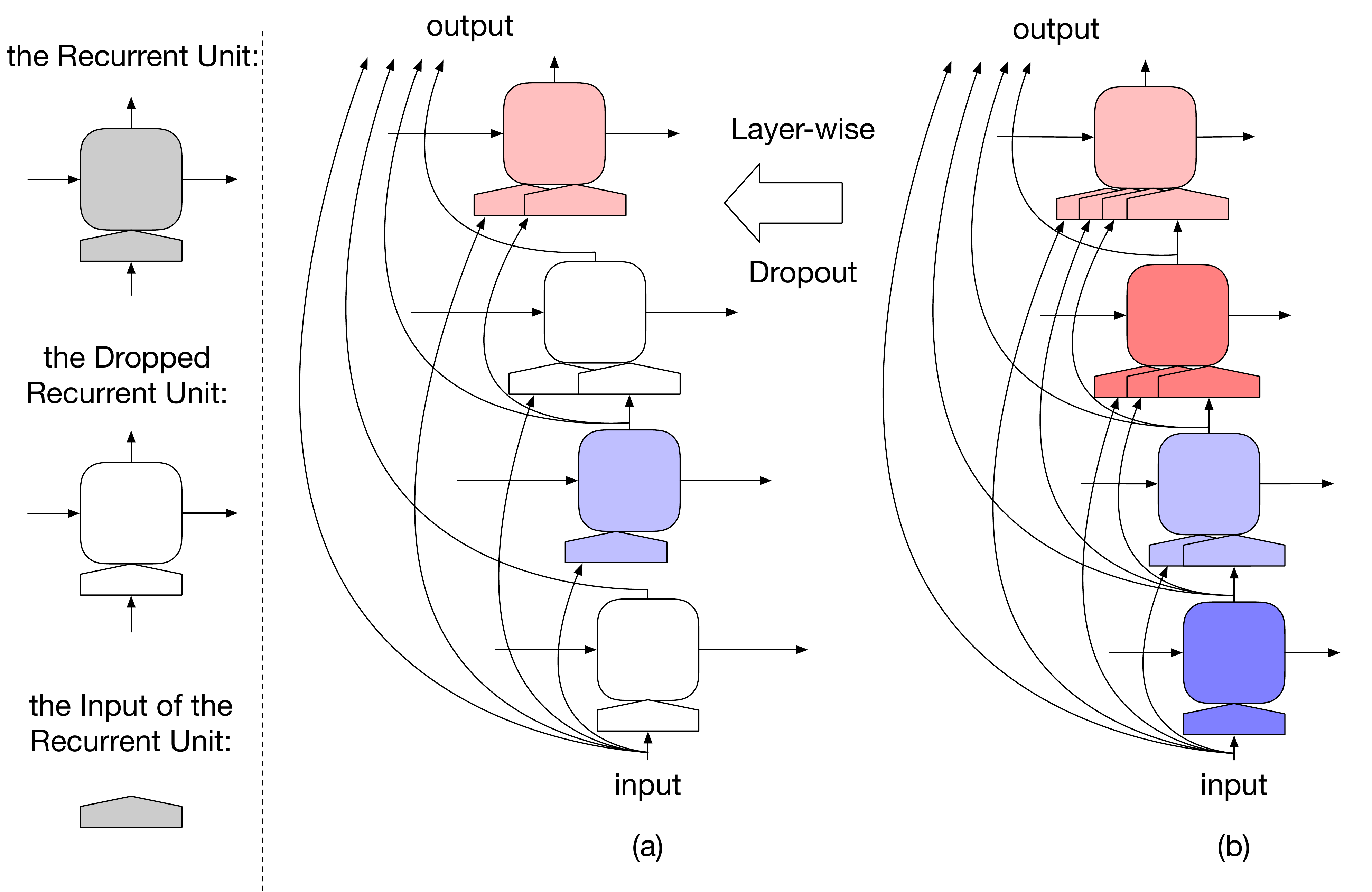}
\vspace{-0.8cm}
  \caption{Layer-wise dropout conducted on a 4-layer densely connected RNN. (a) is the remained RNN. (b) is the original densely connected RNN.}
  \label{fig:Drop}
\vspace{-0.5cm}
\end{figure}

So far, we've customized the regularization term for the layer-wise pruning, which protects the retained connections among layers from shrinking.
After that, we try to further retain the effectiveness of the 
compressed model.
Specifically, we choose to prepare the \plm for the pruned
 inputs, thus making them more robust to 
pruning.

Accordingly, we conduct the training of \plm with a layer-wise dropout.
As in Figure~\ref{fig:Drop}, a random part of layers in the \plm are randomly dropped during each batch.
The outputs of the dropped layers will not be passed to their subsequent recurrent layers, but will be sent to the projection layer (Eqn.~\ref{eqn:h_proj}) for predicting the next word.
In other words, this dropout is only applied to the input of recurrent layers, which aims to imitate the pruned input without totally removing any layers.

\section{Sequence Labeling}

In this section, we will introduce our sequence labeling architecture, which 
is augmented with the contextualized representations.

\subsection{Neural Architecture}
\begin{figure}
  \centering
    \includegraphics[width=0.95\columnwidth]{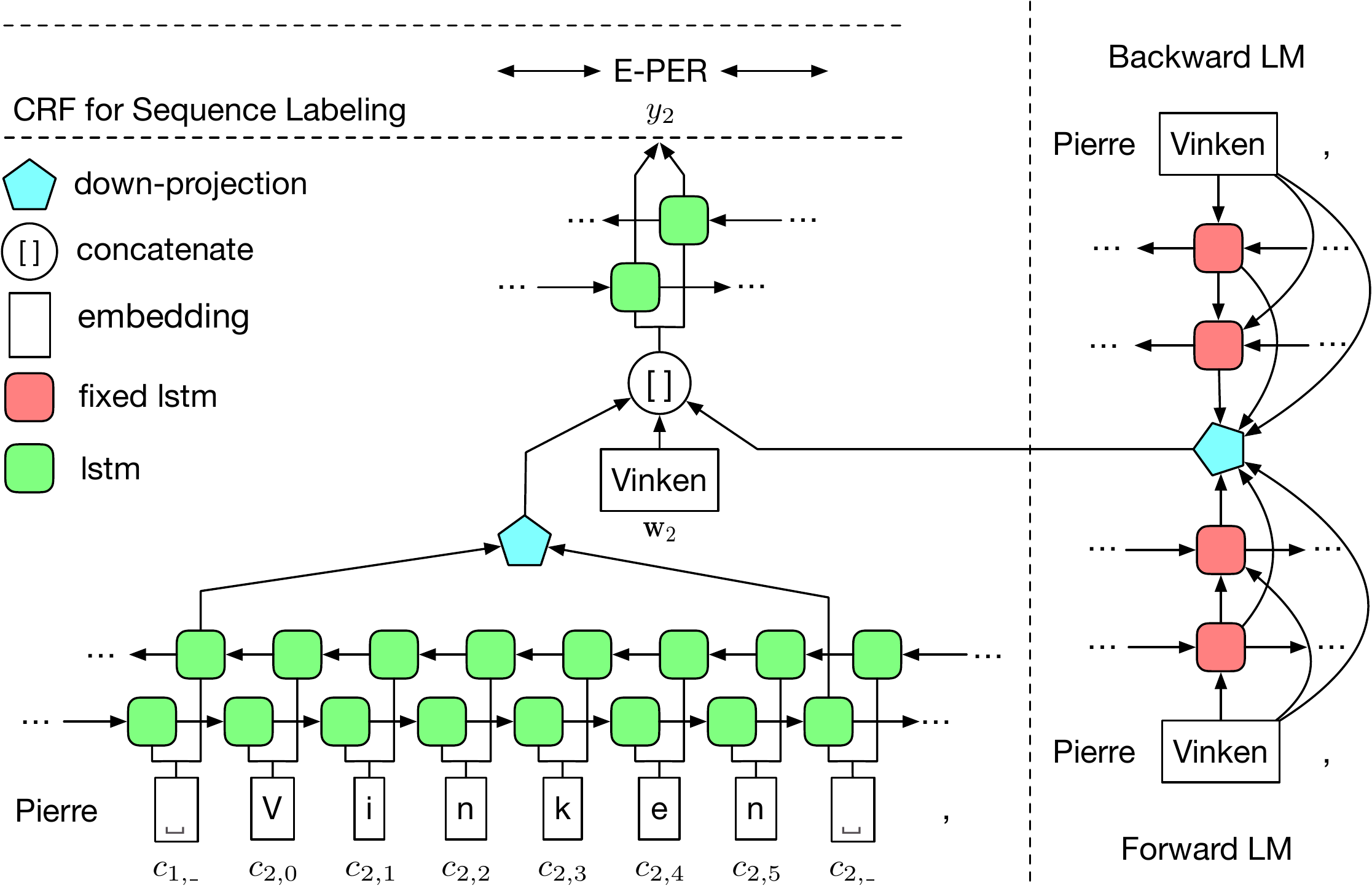}
\vspace{-0.4cm}
  \caption{The proposed sequence labeling architecture with contextualized representations.}
  \label{fig:seq}
\vspace{-0.5cm}
\end{figure}

Following the recent studies~\cite{liu2017empower,kuru2016charner}, we construct the neural architecture as in Fig.~\ref{fig:seq}.
Given the input sequence $\{x_1, x_2, \cdots, x_T\}$, for $t^{th}$ token ($x_t$), we assume its word embedding is $\w_t$, its label is $y_t$, and its character-level input is $\{c_{i, 1}, c_{i, 2}, \cdots, c_{i, \_}\}$, where $c_{i, \_}$ is the space character following $x_t$.

The character-level representations have become the required components for most of the state-of-the-art.
Following the recent study~\cite{liu2017empower},
we employ LSTMs to take the character-level input in a context-aware manner, and mark its output for $x_t$ as $\c_t$.
Similar to the contextualized representation, $\c_t$ usually has more dimensions than $\w_t$.
To integrate them together, we set the output dimension of Eqn.~\ref{eqn:cr} as the dimension of $\w_t$,
and project $\c_t$ to a new space with the same dimension number.
We mark the projected character-level representation as $\c_t^*$.

After projections, these vectors are concatenated as $\v_t = [\c_t^*; \r_t; \w_t], \forall i \in [1, T]$ and further fed into the word-level LSTMs.
We refer to their output as $\U = \{\u_1, \cdots, \u_T\}$.
To ensure the model to predict valid label sequences, we append a first-order conditional random field (CRF) layer to the model~\cite{2016naacl}.
Specifically, the model defines the generation probability of $\y = \{y_1, \cdots, y_T\}$ as 
\begin{equation}
p(\y|\U) = \frac{ \prod_{t=1}^T \phi(y_{t-1}, y_t, \u_t) } { \sum_{\hat{\y} \in \Y(\U)} \prod_{t=1}^T \phi(\hat{y}_{t-1}, \hat{y}_t, \u_t)}
\label{eqn:crf_prob}
\end{equation}
where $\hat{\y} = \{\hat{y}_1, \dots, \hat{y}_T\}$ is a generic label sequence, $\Y(\U)$ is the set of all generic label sequences for $\U$ and $\phi(y_{t-1}, y_t, \u_t)$ is the potential function.
In our model, $\phi(y_{t-1}, y_t, \u_t)$ is defined as $\exp(W_{y_{t}} \u_t + b_{y_{t-1}, y_{t}})$, where $W_{y_{t}}$ and $b_{y_{t-1}, y_{t}}$ are the weight and bias parameters.

\subsection{Model Training and Inference}
We use the following negative log-likelihood as the empirical risk.
\begin{equation}
\mathcal{L} = -\sum_{\U} \log p(\y | \U)
\label{eqn:crf}
\end{equation}
For testing or decoding, we want to find the optimal sequence $\y^*$ that maximizes the likelihood.
\begin{equation}
\y^* = \underset{\y \in \Y(\U)}{{\mbox{argmax}}} \; p(\y | \U)
\label{eqn:decode}
\end{equation}
Although the denominator of Eq.~\ref{eqn:crf_prob} is complicated, we can calculate Eqs.~\ref{eqn:crf} and \ref{eqn:decode} efficiently by the Viterbi algorithm.

For optimization, we decompose it into two steps, i.e., model training and model pruning.

\noindent\textbf{Model training.} 
We set $\lambda_0$ to $0$ and optimize the empirical risk without any regularization, i.e., $\min \mathcal{L}$. 
In this step, we conduct optimization with the stochastic gradient descent with momentum.
Following~\cite{e2018deep}, dropout would be added to both the coupled \plm and the sequence labeling model.

\noindent\textbf{Model pruning.}
We conduct the pruning based on the checkpoint which has the best performance on the development set during the model training.
We set $\lambda_0$ to non-zero values and optimize $\min \mathcal{L} + \lambda_0 \mathcal{R}_3$ by the projected gradient descent with momentum. 
Any layer $i$ with $z_i = 0$ would be deleted in the final model to complete the pruning.
To get a better stability, dropout is only added to the sequence labeling model.

%% file: 3exp.tex

\begin{table*}[ht!]
\begin{center}
\begin{tabularx}{\textwidth}{c *{6}{Y}}
\toprule
\multirow{2}{*}{Network} & \multirow{2}{*}{Ind. \# } & \multirow{2}{*}{Hid. \#} & \multirow{2}{*}{Layer \#} & \multicolumn{2}{c}{Param.\# ($\cdot 10^7$)} & \multirow{2}{*}{PPL} \\
\cmidrule(lr){5-6}
 & & & & RNN & Others & \\
\midrule
\midrule
8192-1024~\cite{jozefowicz2016exploring} 
 & 1 & 8192 & 2 & 15.1$^\sharp$ & 163$^\sharp$ & 30.6 \\
\midrule
CNN-8192-1024~\cite{jozefowicz2016exploring} 
 & 2 & 8192 & 2 & 15.1$^\sharp$ & 89$^\sharp$ & 30.0 \\
\midrule
CNN-4096-512~\cite{e2018deep}
 & 3 & 4096 & 2 & 3.8$^\sharp$ & 40.6$^\sharp$ & 39.7 \\
\midrule
2048-512~\cite{peters2017semi}
 & 4 & 2048 & 1 & 0.9$^\sharp$ & 40.6$^\sharp$ & 47.50 \\
\midrule
2048-Adaptive~\cite{grave2016efficient} 
 & 5 & 2048 & 2 & 5.2$^\dagger$ & 26.5$^\dagger$ & 39.8 \\
\midrule
\multirow{2}{*}{vanilla LSTM} 
 & 6 & 2048 & 2 & 5.3$^\dagger$ & 25.6$^\dagger$ & 40.27\\
 & 7 & 1600 & 2 & 3.2$^\dagger$ & 24.2$^\dagger$ & 48.85\\
\midrule
\midrule
\our without Layer-wise Dropout
 & 8 & 300 & 10 & 2.3$^\dagger$ & 24.2$^\dagger$ & 45.14\\
\midrule
\our with Layer-wise Dropout
 & 9 & 300 & 10 & 2.3$^\dagger$ & 24.2$^\dagger$ & 50.06\\
\bottomrule
\end{tabularx}
\end{center}
\vspace{-0.4cm}
\caption{Performance comparison of language models. Models marked with$^\dagger$ adopted adaptive softmax and the vanilla LSTMs, which has less softmax parameters. Models marked with$^\sharp$ employed sampled softmax LSTMs w. projection, which results in less RNN parameters w.r.t. the size of hidden states.}
\vspace{-0.5cm}
\label{tab:lm}
\end{table*}

\section{Experiments}
\label{sec:exp}

We will first discuss the capability of the \our as language models, then explore the effectiveness of its contextualized representations.

\subsection{Language Modeling}

For comparison, we conducted experiments on the one billion word benchmark dataset~\cite{41880} with both \our (with 1,600 dimensional projection) and the vanilla stacked-LSTM.
Both kinds of models use word embedding (random initialized) of 300 dimension as input and use the adaptive softmax (with default setting) as an approximation of the full softmax.
Additionally, as preprocessing, we replace all tokens occurring equal or less than 3 times with as UNK, which shrinks the dictionary from 0.79M to 0.64M. 

The optimization is performed by the Adam algorithm~\cite{kingma2014adam}, the gradient is clipped at $5.0$ and the learning rate is set to start from $0.001$.
The layer-wise dropout ratio is set to $0.5$, the RNNs are unrolled for 20 steps without resetting the LSTM states, and the batch size is set to 128.
Their performances are summarized in Table~\ref{tab:lm}, together with several LMs used in our sequence labeling baselines.
For models without official reported parameter numbers, we estimate their values (marked with$^\dagger$) by assuming they adopted the vanilla LSTM. 
Note that, for models 3, 5, 6, 7, 8, and 9, PPL refers to the averaged perplexity of the forward and the backward LMs.

We can observe that, for those models taking word embedding as the input, embedding composes the vast majority of model parameters.
However, embedding can be embodied as a ``sparse'' layer which is computationally efficient.
Instead, the intense calculations are conducted in RNN layers and softmax layer for language modeling, or RNN layers for contextualized representations.
At the same time, comparing the model 8192-1024 and CNN-8192-1024, their only difference is the input method.
Instead of taking word embedding as the input, CNN-8192-1024 utilizes CNN to compose word representation from the character-level input.
Despite the greatly reduced parameter number, the perplexity of the resulting models remains almost unchanged.
Since replacing embedding layer with CNN would make the training slower, we only conduct experiments with models taking word embedding as the input.

Comparing \our with other baselines, we think it achieves satisfactory performance with regard to the size of hidden states.
It demonstrates the \our's capability of capturing the underlying structure of natural language.
Meanwhile, we find that the layer-wise dropout makes it harder to train \our and its resulting model achieves less competitive results.
However, as would be discussed in the next section, layer-wise dropout allows the resulting model to generate better contextualized representations and be more robust to pruning, even with a higher perplexity.

\subsection{Sequence Labeling}

Following 
TagLM~\cite{peters2017semi}, we evaluate our 
methods
in two benchmark datasets, the CoNLL03 NER task~\cite{tjong2003introduction} and the CoNLL00 Chunking task~\cite{tjong2000introduction}.

\noindent\textbf{CoNLL03 NER} has four entity types
and includes the standard training, development and test sets.

\noindent \textbf{CoNLL00 chunking} defines eleven syntactic chunk types (e.g., \texttt{NP} and \texttt{VP}) in addition to \texttt{Other}. Since it only includes training and test sets, we sampled 1000 sentences from training set as a held-out development set~\cite{peters2017semi}.

In both cases, we use the BIOES labeling scheme~\cite{ratinov2009design} and use the micro-averaged $F_1$ as the evaluation metric. 
Based on the analysis conducted in the development set, we set $\lambda_0 = 0.05$ for the NER task, and $\lambda_0 = 0.5$ for the Chunking task.
As discussed before, we conduct optimization with the stochastic gradient descent with momentum.
We set the batch size, the momentum, and the learning rate to $10$, $0.9$, and $\eta_t = \frac{\eta_0}{1+\rho t}$ respectively. Here, $\eta_0 = 0.015$ is the initial learning rate and $\rho = 0.05$ is the decay ratio. 
Dropout is applied in our model, and its ratio is set to $0.5$. 
For a better stability, we use gradient clipping of $5.0$.
Furthermore, we employ the early stopping in the development set and report averaged score across five different runs.

Regarding the network structure, we use the 30-dimension character-level embedding.
Both character-level and word-level RNNs are set to one-layer LSTMs with 150-dimension hidden states in each direction.
The GloVe 100-dimension pre-trained word embedding\footnote{
\fontsize{8pt}{8pt} 
\url{https://nlp.stanford.edu/projects/glove/}} is used as the initialization of word embedding $\w_t$, and will be fine-tuned during the training.
The layer selection variables $z_i$ are initialized as $1$, remained unchanged during the model training and only be updated during the model pruning.
All other variables are randomly initialized~\cite{glorot2010understanding}.

\begin{table}[t]
\begin{center}
\begin{tabularx}{\columnwidth}{r c *{2}{Y}}
\toprule
Network & Avg. & \#FLOPs & $F_1$ score \\
(\small \plm Ind.\#) & ppl & ($\cdot 10^6$)  & (avg{\small $\pm$}std) \\
\midrule
\midrule
\nolm (/) & / & 3 & 94.42{\small $\pm$}0.08 \\
\midrule
\nel (6) & 40.27 & 108 & 96.19{\small $\pm$}0.07 \\
\midrule
\nel (7) & 48.85 & 68 & 95.86{\small $\pm$}0.04 \\
\midrule
\midrule
\our$^*$ (8) & 45.14 & 51 & 96.01{\small $\pm$}0.07\\
\midrule
\our$^*$ (9) & 50.06 & 51 & 96.05{\small $\pm$}0.08\\
\midrule
\multirow{2}{*}{\our (8)}
 & {\small origin} & 51 & 96.13\\
 & {\small pruned} &  13 & 95.46{\small $\pm$}0.18 \\
\midrule
\multirow{2}{*}{\our (9)}
 & {\small origin} & 51 & 96.15\\
 & {\small pruned} & 10 & 95.66{\small $\pm$}0.04 \\
\bottomrule
\end{tabularx}
\end{center}
\vspace{-0.3cm}
\caption{Performance comparisons in the CoNLL00 Chunking task. LD-Net maked with $^*$ are trained without pruning (layer selection).}
\label{tab:np}
\vspace{-0.6cm}
\end{table}

\begin{table}
\begin{center}
\begin{tabularx}{\columnwidth}{r c *{2}{Y}}
\toprule
Network & Avg. & \#FLOPs & $F_1$ score \\
(\small \plm Ind.\#) & ppl & ($\cdot 10^6$)  & (avg{\small $\pm$}std) \\
\midrule
\midrule
\nolm (/) & / & 3 & 90.78{\small $\pm$}0.24 \\
\midrule
\oel (3) & 39.70 & 79$^\sharp$ & 92.22{\small $\pm$}0.10 \\
\midrule
TagLM (4) & 47.50 & 22$^\sharp$ & 91.62{\small $\pm$}0.23 \\
\midrule
\nel (6) & 40.27 & 108 & 91.99{\small $\pm$}0.24 \\
\midrule
\nel (7) & 48.85 & 68 & 91.54{\small $\pm$}0.10 \\
\midrule
\midrule
\our$^*$ (8) & 45.14 & 98 & 91.76{\small $\pm$}0.18\\
\midrule
\our$^*$ (9) & 50.06 & 98 & 91.86{\small $\pm$}0.15 \\
\midrule
\multirow{2}{*}{\our (8)}
 & {\small origin} & 51 & 91.95\\
 & {\small pruned} & 5 & 91.55{\small $\pm$}0.06 \\
\midrule
\multirow{2}{*}{\our (9)}
 & {\small origin} & 51 & 92.03\\
 & {\small pruned} & 5 & 91.84{\small $\pm$}0.14 \\
\bottomrule
\end{tabularx}
\end{center}
\vspace{-0.4cm}
\caption{Performance comparison in the CoNLL03 NER task. Models marked with$^\dagger$ employed LSTMs with projection, which is more efficient than the vanilla LSTMs. LD-Net maked with $^*$ are trained without pruning (layer selection).}
\label{tab:ner}
\vspace{-0.4cm}
\end{table}

\begin{table*}[t]
\begin{center}
\begin{tabularx}{\textwidth}{r c *{5}{Y}}
\toprule
\multirow{2}{*}{Network (\plm Ind.\#)}  & \multirow{2}{*}{FLOPs}    & \multirow{2}{*}{Batch size}   & \multirow{2}{*}{Peak RAM} & \multirow{2}{*}{Time (s)}   & \multicolumn{2}{c}{Speed} \\ \cline{6-7}
                                        &                           &                               &                           &                               & $10^3$words/s    & $10^3$sents/s        \\ \midrule \midrule
\nel (6)                                & 108                       & 200                           & 8Gb                       & 32.88                         & 22       & 0.4         \\ \midrule
\our (9, origin)                        & 51                        & 80                            & 8Gb                       & 25.68                         & 26       & 0.5         \\ \midrule
\multirow{2}{*}{\our (9, pruned)}       & \multirow{2}{*}{5}        & 80                            & 4Gb                       & 6.90                          & 98       & 2.0         \\ \cline{3-7}
                                        &                           & 500                           & 8Gb                       & 4.86 (\textbf{5X})            & 166 (\textbf{6X}) & 2.9 (\textbf{5X})     \\
\bottomrule
\end{tabularx}
\end{center}
\vspace{-0.4cm}
\caption{Speed comparison in the CoNLL03 NER task. We can observe that LD-Net (9, pruned) achieved about 5 times speed up on the wall-clock time over LD-Net (9, origin).}
\label{tab:speed}
\vspace{-0.5cm}
\end{table*}

\noindent \textbf{Compared methods.}
The first baseline, referred as \nolm, is our sequence labeling model without the contextualized representations, i.e., calculating $\v_t$ as $[\c_t^*; \w_t]$ instead of $[\c_t^*; \r_t; \w_t]$.
Besides, ELMo~\cite{e2018deep} is the major baseline.
To make comparison more fair, we implemented the ELMo model and use it to calculate the $\r_t$ in Eqn.~\ref{eqn:cr} instead of $[\h_t, \h_t^r]$.
Results of re-implemented models are referred with \nel ($\lambda$ is set to the recommended value, $0.1$) and the results reported in its original paper are referred with \oel.
Additionally, since TagLM~\cite{peters2017semi} with one-layer NNs can be viewed as a special case of ELMo, we also include 
its results.

\noindent \textbf{Sequence labeling results.}
Table~\ref{tab:np} and ~\ref{tab:ner} summarizes the results of \our and baselines.
Besides the $F_1$ score and averaged perplexity, we also estimate FLOPs (i.e., the number of floating-point multiplication-adds) for the efficiency evaluation.
Since our model takes both word-level and character-level inputs, we estimated the FLOPs value for a word-level input with $4.39$ character-level inputs, while $4.39$ is the averaged length of words in the CoNLL03 dataset.

Before the model pruning, \our achieves a 96.05$\pm$0.08 $F_1$ score in the CoNLL00 Chunking task, yielding nearly 30\% error reductions over the NoLM baseline.
Also, it scores 91.86$\pm$0.15 $F_1$ in the CoNLL03 NER task with over 10\% error reductions.
Similar to the language modeling, we observe that the most complicated models achieve the best perplexity and provide the most improvements in the target task.
Still, considering the number of model parameters and the resulting perplexity, our model demonstrates its effectiveness in generating contextualized representations.
For example, comparing to our methods, \nel (7) leverages \plm with the similar perplexity and parameter number, but cannot get the same improvements with our method on both datasets.

Actually, contextualized representations have strong connections with the skip-thought vectors~\cite{kiros2015skip}.
Skip-thought models try to embed sentences and are trained by predicting the previous and afterwards sentences.
Similarly, LMs encode a specific context as the hidden states of RNNs, and use them to predict future contexts.
Specifically, we recognize the cell states of LSTMs are more like to be the sentence embedding~\cite{radford2017learning}, since they are only passed to the next time stamps.
At the same time, because the hidden states would be passed to other layers, we think they are more like to be the token representations capturing necessary signals for predicting the next word or updating context representations\footnote{We tried to combine the cell states with the hidden states to construct the contextualized representations by concatenation or weighted average, but failed to get better performance. We think it implies that ELMo works as token representations instead of sentence representations}.
Hence, LD-Net should be more effective then ELMo, as concatenating could preserve all extracted signals while weighted average might cause information loss.

Although the layer-wise dropout makes the model harder to train, their resulting \plm generate better contextualized representations, even without the same perplexity.
Also, as discussed in~\cite{e2018deep,peters2017semi}, the performance of the contextualized representation can be further improved by training larger models or using the CNN to represent words.

For the pruning, we started from the model with the best performance on the development set (referred with ``origin''), and refer the performances of pruned models with ``pruned'' in Table~\ref{tab:np} and ~\ref{tab:ner}.
Essentially, we can observe the pruned models get rid of the vast majority of calculation while still retaining a significant improvement.
We will discuss more on the pruned models in Sec.~\ref{subsec:cs}.

\subsection{Speed Up Measurements}
\label{subsec:speed}

We use FLOPs for measuring the inference efficiency as it reflects the time complexity~\cite{han2015learning}, and thus is independent of specific implementations.
For models with the same structure, improvements in FLOPs would result in monotonically decreasing inference time. 
However, it may not reflect the actual efficiency of models due to the model differences in parallelism.
Accordingly, we also tested wall-clock speeds of our implementations.

Our implementations are based on the PyTorch 0.3.1\footnote{\url{http://pytorch.org/}}, and all experiments are conducted on the CoNLL03 dataset with the Nvidia GTX 1080 GPU.
Specifically, due to the limited size of CoNLL03 test set, we measure such speeds on the training set.
As in Table~\ref{tab:speed}, we can observe that, the pruned model achieved about 5 times speed up.
Although there is still a large margin between the actual speed-up and the FLOPs speed-up, we think the resulting decode speed (166K words/s) is sufficient for most real-world applications. 

\subsection{Case Studies}
\label{subsec:cs}

\begin{figure}
  \centering
    \includegraphics[width=\columnwidth]{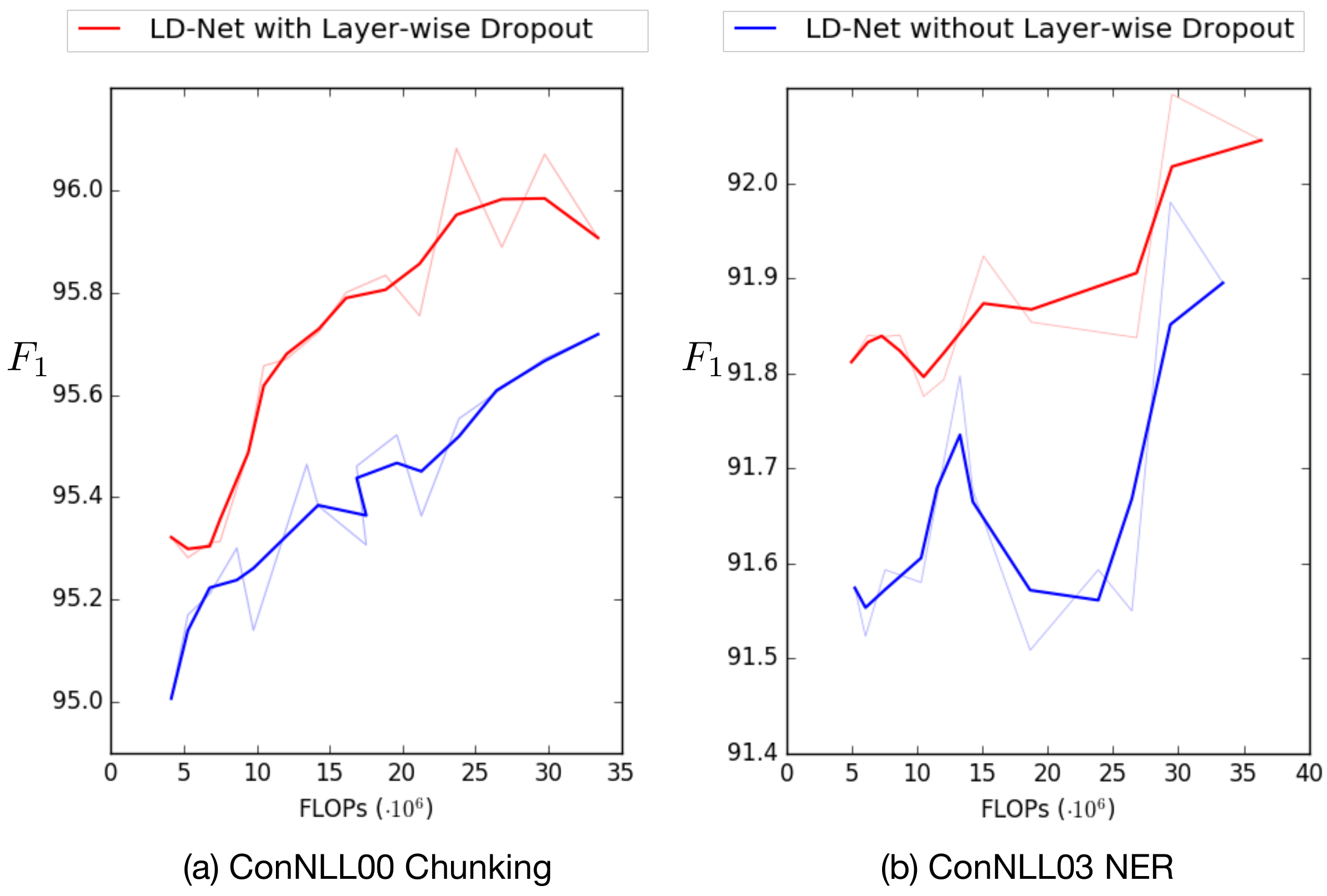}
\vspace{-0.8cm}
  \caption{The performance of pruned models in two tasks w.r.t. their efficiency (FLOPs).}
  \label{fig:curve}
\vspace{-0.5cm}
\end{figure}

\noindent \textbf{Effect of the pruning ratio.}
To explore the effect of the pruning ratio, we adjust $\lambda_1$ and visualize the performance of pruned models v.s. their FLOPs \# in Fig~\ref{fig:curve}.
We can observe that \our outperforms its variants 
and demonstrates its effectiveness.

As the pruning ratio becoming larger, we can observe the performance of \our first increases a little, then starts dropping. 
Besides, in the CoNLL03 NER task, \plm can be pruned to a relatively small size without much loss of efficiency.
As in Table~\ref{tab:ner}, we can observe that, after pruning over 90\% calculations, the error of the resulting model only increases about 2\%, yielding a competitive performance.
As for the CoNLL00 Chunking task, the performance of \our decays in a faster rate than that in the NER task.
For example, after pruning over 80\% calculations, the error of the resulting model increases about 13\%.
Considering the fact that this corpus is only half the size of the CoNLL03 NER dataset, we can expect the resulting models have more dependencies on the \plm.
Still, the pruned model achieves a 25\% error reduction over the NoLM baseline.

\begin{figure}
  \centering
    \includegraphics[width=\columnwidth]{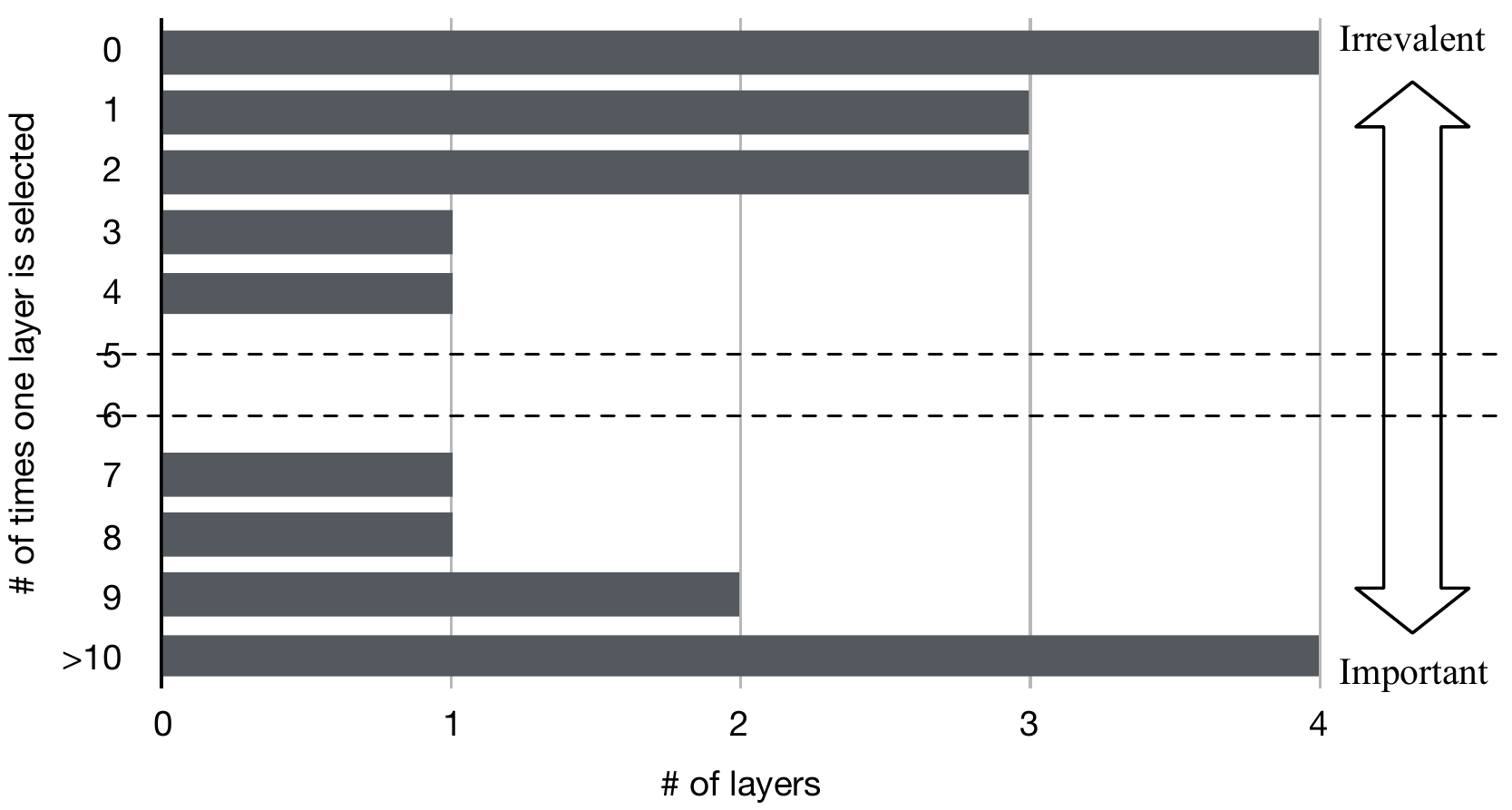}
\vspace{-0.8cm}
  \caption{The performance of pruned models in two tasks w.r.t. their efficiency (FLOPs).}
  \label{fig:bar}
\vspace{-0.5cm}
\end{figure}

\noindent \textbf{Layer selection pattern.}
We further studied the layer selection patterns.
Specifically, we use the same setting of LD-Net (9) in Table~\ref{tab:ner}, conduct model pruning using for 50 times, and summarize the statics in Figure~\ref{fig:bar}.
We can observe that network layers formulate two clear clusters, one is likely to be preserved during the selection, and the other is likely to be pruned. 
This is consistent with our intuition that some layers are more important than others and the layer selection algorithm would pick up layers meaningfully.

However, there is some randomness in the selection result.
We conjugate that large networks trained with dropout can be viewed as a ensemble of small sub-networks~\cite{hara2016analysis}, also there would be several sub-networks having the similar function.
Accordingly, we think the randomness mainly comes from such redundancy.

\noindent \textbf{Effectiveness of model pruning.}
\citet{zhu2017prune} observed pruned large models consistently outperform small models on various tasks (including LM).
These observations are consistent with our experiments. For example, LD-Net achieves 91.84 after pruning on the CoNLL03 dataset. 
It outperforms TagLM (4) and R-ELMo (7), whose performances are $91.62$ and $91.54$. 
Besides, we trained small LMs of the same size as the pruned LMs (1-layer densely connected LSTMs). 
Its perplexity is 69 and its performance on the CoNLL03 dataset is $91.55\pm0.19$.

%% file: 4related.tex

\section{Related Work}

\noindent \textbf{Sequence labeling.}
Linguistic sequence labeling is one of the fundamental tasks in NLP, encompassing various applications including POS tagging, chunking, and NER.
Many attempts have been made to 
 conduct end-to-end learning and build reliable models without handcrafted features~\cite{Chiu2016NamedER,2016naacl,ma-hovy:2016:P16-1}.

\noindent \textbf{Language modeling.}
Language modeling is a core task in NLP.
Many attempts have been paid to develop better neural language models~\cite{zilly2016recurrent,inan2016tying,godin2017improving,Melis2017OnTS}.
Specifically, with extensive corpora, language models can be well trained to generate high-quality sentences from scratch~\cite{jozefowicz2016exploring,grave2016efficient,li2017slim,shazeer2017outrageously}.
Meanwhile, initial attempts have been made to improve the performance of other tasks with these methods.
Some methods treat the language modeling as an additional supervision, and conduct co-training for knowledge transfer~\cite{dai2015semi, liu2017empower,rei2017semi}.
Others, including this paper, aim to construct additional features (referred as contextualized representations) with the pre-trained language models~\cite{peters2017semi,e2018deep}.

\noindent \textbf{Neural Network Acceleration.}
There are mainly three kinds of NN acceleration methods, i.e., prune network into smaller sizes~\cite{han2015learning,wen2016learning}, converting float operation into customized low precision arithmetic~\cite{hubara2016quantized,courbariaux2016binarized}, and using shallower networks to mimic the output of deeper ones~\cite{hinton2015distilling,romero2014fitnets}.
However, most of them require costly retraining.

%% file: 5con.tex

\section{Conclusion}

Here, we proposed \our, a novel framework for efficient contextualized representation. 
As demonstrated on two benchmarks, it can conduct the layer-wise pruning 
for a specific task.
Moreover, it requires neither the gradient oracle of \plm nor the costly retraining.
In the future, we plan to apply \our to other
applications.

%% file: 6akn.tex
\section*{Acknowledgments}
\label{sect:ack}

We thank Junliang Guo and all reviewers for their constructive comments.
Research was sponsored by the Army Research Laboratory and was accomplished under Cooperative Agreement Number W911NF-09-2-0053 (the ARL Network Science CTA). The views and conclusions in this document are those of the authors and should not be interpreted as representing the official policies, either expressed or implied, of the Army Research Laboratory or the U.S. Government. The U.S. Government is authorized to reproduce and distribute reprints for Government purposes notwithstanding any copyright notation here on.